\documentclass{article}

\usepackage[nonatbib,
final 
]{neurips_2024}

\usepackage[utf8]{inputenc} 
\usepackage[T1]{fontenc}    
\usepackage[hidelinks]{hyperref}       
\usepackage{url}            
\usepackage{booktabs}       
\usepackage{amsfonts}       
\usepackage{nicefrac}       
\usepackage{microtype}      
\usepackage{xcolor}         

\usepackage{amsmath}

\usepackage{appendix}
\usepackage{graphicx}

\usepackage{subcaption} 

\usepackage[
backend=biber,
sorting=none,
style=numeric,
]{biblatex} 
\title{The Intersectionality Problem for Algorithmic Fairness}

\author{%
  Johannes Himmelreich\\ 
  Syracuse University\\
  \texttt{jrhimmel@syr.edu} \\
  \And
Arbie Hsu\\
University of San Francisco\\
  \texttt{whsu10@dons.usfca.edu} \\
  \AND
Kristian Lum\\
University of Chicago\\
  \texttt{kristianl@uchicago.edu} \\
  \And
Ellen Veomett\\
University of San Francisco\\
  \texttt{eveomett@usfca.edu} \\
}

\begin{document}
\maketitle

\begin{abstract} 
A yet unmet challenge in algorithmic fairness is the problem of intersectionality, that is, achieving fairness across the intersection of multiple groups---and \emph{verifying} that such fairness has been attained. Because intersectional groups tend to be small, verifying whether a model is fair raises statistical as well as moral-methodological challenges. This paper (1) elucidates the problem of intersectionality in algorithmic fairness, (2) develops desiderata to clarify the challenges underlying the problem and guide the search for potential solutions, (3) illustrates the desiderata and potential solutions by sketching a proposal using simple hypothesis testing, and (4) evaluates, partly empirically, this proposal against the proposed desiderata.
\end{abstract}

\section{Introduction}
\emph{That} intersectionality matters is a point of consensus in the algorithmic fairness literature. A model's performance might be much worse for women of color than for women and people of color considered separately \cite{GenderShades}. In this paper, we elucidate a problem that intersectionality raises for algorithmic fairness in practice: Because data on intersectional groups is often severely limited, \emph{verifying} that algorithmic fairness---under various definitions thereof---has been attained is difficult. Although this problem is recognized in the literature \cite{Kearns_FairnessGerrymandering, 9101635, AuditingAchievingIntersectional, MolinaLoiseauBoundingApproximating}, its challenges do not appear to be fully appreciated and many existing contributions violate minimal moral or methodological desiderata.

Our contribution is fourfold: We (1) elucidate the problem of intersectionality in algorithmic fairness, and (2) develop desiderata to clarify the challenges that underlie the problem of intersectionality and to guide the search for potential solutions. Moreover, we (3) illustrate the desiderata and potential solutions by presenting a statistical setup that uses simple hypothesis testing, and (4) evaluate this proposal, partly empirically, in light of the desiderata. 

Our larger aim is to advance the literature on algorithmic fairness more broadly. The approach that we propose in response to the problem of intersectionality differs fundamentally from the typical way of ``measuring'' algorithmic fairness.\footnote{We use fairness ``measure,'' ``metric'' and their cognates with two caveats. First, the problem is one of estimation not measurement. Second, fairness metrics are \emph{meta-metrics} since they aggregate a higher-dimensional vector of model performance into a lower-dimensional summary \cite{lum_-biasing_2022}.}  We hence advance the debate, by pointing out possibilities of approaching fairness differently: as accounting for uncertainty (instead of concentrating on point estimates) and as a matter of sufficiency (instead of equality). 

\section{Preliminaries}
\subsection{Algorithmic Fairness}

In the literature on algorithmic fairness, ``fairness'' is typically defined as model performance (such as accuracy or false positive rate) that is roughly equal across all relevant groups. Many versions of algorithmic fairness consider fairness to have been achieved if 
\begin{equation}\label{eq:typical}
    |m(G) - m(\cdot)| < \epsilon \quad \text{ for some small } \epsilon, \ \forall \, G
\end{equation}

Where $G$ denotes a subgroup of the population, $m(G)$ a model's performance (however understood) on only the subset of the data that belongs to group $G$, and $m(\cdot)$ the model's performance calculated across the entire dataset, irrespective of group membership. 
Membership in $G$ typically corresponds to a sensitive or protected attribute such as race, sex, age, disability or marital status but $G$ may also be defined intersectionally as a \emph{combination} of such attributes. 

Equation \eqref{eq:typical} generalizes a large family of definitions or---when aggregating $|m(G) - m(\cdot)|$ for all groups---metrics of fairness. We thus take \eqref{eq:typical} to represent the \emph{typical} way of understanding algorithmic fairness. This typical way of understanding fairness faces the problem of intersectionality.

\subsection{The Problem of Intersectionality}\label{sec:intersectionality}

As the number of attributes that define subgroups grows, the amount of data available for each subgroup shrinks rapidly. After all, the number of subgroups grows \emph{exponentially} with the number of protected attributes: For $n$ binary attributes, there are $2^n$ intersectional groups. This, in turn, entails a data problem: When social identities are constituted by intersections of increasingly many attributes, and when these constituting attributes are not just binary, the data within each of the intersections can become very small. In Europe, where discrimination is highly intersectional and fairness audits are encouraged by legislation,\footnote{Recital 49 of the EU Artificial Intelligence Act \cite{EUAI} encourages ``the development of benchmarks and measurement methodologies for AI systems'' \cite{Recital49}. Yet the statistical problems of intersectional fairness are, in some way, greater in Europe. Since nationality groups are already comparatively small, intersectional groups are even smaller subgroups within already small nationality groups. For example, Hungarian Roma face discrimination in the housing market, Maghrebi French in the labor market, whereas people of African descent in England and Wales face discrimination in the criminal justice system \cite{IntersectionalityEurope}.} fairness audits may need to account for several thousand subgroups.\footnote{Assuming 3 binary attributes (e.g., non-white, cis-gender, same-sex orientation) 1 three-valued attribute (e.g., gender as `male,' `female,' and `neither'), 9 different ethnic backgrounds (e.g., Roma, Chinese, Turkish), and 12 different nationalities or localities (e.g., Hungarian, German, French), yields 2,592 intersectional subgroups. And this number may be conservative since the number of discernible ethnic groups is larger than 9, of nationalities is larger than 12, and the legally protected attribute of age is not even included.} And because gathering the data necessary for fairness audits is typically costly---e.g., the ``ground truth'' needs to be established to assess whether a prediction is correct---such data tend to be scarce to begin with.

In short, the intersectionality problem of algorithmic fairness is a problem of statistical uncertainty due to small data and, subsequently, raises problems for how ``fairness'' is typically defined. 

Intersectionality renders fairness metrics, as they are typically defined, meaningless. These metrics, such as \eqref{eq:typical}, rely on point estimates of model performance (e.g., whether this performance is roughly the same for all groups). But point estimates become nonsensical with small data \cite{Kearns_FairnessGerrymandering}.\footnote{For example, in binary classification, an individual prediction is either 1 or 0; and the model accuracy for each singleton group is thus either 1 or 0.}
The challenge posed by intersectionality for algorithmic fairness is to define a fairness metric that provides meaningful estimates of fairness even when groups are very small and audit data are sparse.

Our discussion hence adds to the existing technical and critical objections against (intersectional) algorithmic fairness \cite{corbett-davies_measure_2024,kong_are_2022}, acknowledging that a commitment to intersectionality and fairness likely requires a broader set of actions than estimating certain properties of models \cite{wang_towards_2022,suresh_towards_2022,klumbyte_critical_2022}.

\section{Existing Work}
\label{sec:existinG_work}

Various statistical methods have been proposed for intersectionality in algorithmic fairness.

\subsection{Kearns et al.}
An early identification and statement of the problem of intersectional fairness arising from small groups is due to Kearns et al.\ \cite{Kearns_FairnessGerrymandering}. The approach of Kearns et al.\ involves an audit algorithm that learns to classify models as fair or unfair instead of defining a fairness metric. The process of learning this audit algorithm is subject to a fairness constraint that is weighted depending on the proportion of the population belonging to a particular group $G$. 

Kearns et al.\ define $\alpha(G)  = Pr(G)$  and reformulate fairness in \eqref{eq:typical} as
\begin{equation}\label{eq:Kearns}
    \alpha(G) |m(G) - m(\cdot)| < \epsilon \quad  \forall \, G
\end{equation}

Essentially, the addition of $\alpha(G)$ relaxes the original fairness metric of \eqref{eq:typical} depending on the proportion of $G$ as a share of the overall population. The smaller $G$ is, the more the condition is relaxed. As Kearns et al.\ explain, this addition is necessary to enable statistical estimation, given the increasing statistical uncertainty with decreasing group size. We discuss the implications in Section \ref{sec:desiderata}, and give the results of an empirical study regarding this formulation in Appendix \ref{appendix:Kearns}.

\subsection{Foulds et al. and Morina et al.}
Foulds et al.\ \cite{9101635} provide an alternative approach based on ratios of model performance metrics. An expanded version of which is, in turn, given by Morina et al.\ \cite{AuditingAchievingIntersectional}.\footnote{We note that Foulds et al.\ (the same group as in \cite{9101635}) also study the usage of Bayesian modeling to more accurately measure fairness metrics than point estimates \cite{doi:10.1137/1.9781611976236.48}.  While these Bayesian models for measuring fairness metrics may give more accurate estimates than point estimates, they do not allow for for the same kind of statistical analysis and ethical evaluation as a confidence interval (that we propose in Sections \ref{sec:desiderata} and \ref{sec:Our_models}).}

These definitions require that the ratio of some metric value between two groups be within a fixed interval.  For example, suppose $m(G)$ measures the true positive rate (TPR) for subgroup $G$.  Then the $\epsilon$-differential intersectional definition of TPR parity (equal opportunity) given in \cite{AuditingAchievingIntersectional} is that
\begin{equation}\label{eq:fairness_ratio}
    e^{-\epsilon} \leq \frac{m(G)}{m(G')} \leq e^\epsilon \quad  \forall \, G, G'
\end{equation}
Morina et al.\ \cite{AuditingAchievingIntersectional} note that $\epsilon = 0$ corresponds to ``perfect fairness'' ($m(G) = m(G')$). 

\subsection{Molina and Loiseau}
Molina and Loiseau use a statistical approach to addressing intersectionality and fairness \cite{MolinaLoiseauBoundingApproximating}. They call a classifier $(\epsilon, \delta)$-probably intersectionally fair if ``the expected number of people that faces a discrimination more than $\epsilon$ is less than $n\delta$'' ($n$ is the population size).\footnote{Molina and Loiseau moreover highlight the issue of estimating fairness of a model on subgroups for whom the set of predicted values on that subgroup is a proper subset of the set of all predicted values.  This becomes an issue because they use a ratio similar to that in Equation \eqref{eq:fairness_ratio}, which is undefined if $m(g')=0$ for some group $g'$.  Our models do not suffer from this issue; however, extremely tiny subgroups do come with their own statistical uncertainty issues, as we highlight in Section \ref{sec:Our_models}.}

\subsection{Cherian and Candès}

Cherian and Candès \cite{CherianCandes} address fairness auditing for many subpopulations within the framework of hypothesis testing, as we do here.  They use a bootstrap process to provide statistical performance bounds for many subpopulations at once. Our addition to this study is the illumination and discussion of desiderata (in Section \ref{sec:desiderata}), a clear description of how one can design fairness metrics using hypothesis testing (Section \ref{sec:Our_models}), and an empirical study showing that these metrics encourage (rather than discourage) the gathering of additional data to improve model performance (Section \ref{sec:analysis}).

\subsection{Khan et al.\ and Agrawal et al.}

Khan et al.\ \cite{khan2023fairnessstabilityestimatorvariance} consider metrics of fairness, accuracy, and variance for model estimators.  They empirically show that there tends to be a tradeoff between these three values.  In a similar vein, Agrawal et al.\ study debiasing methods, and in doing so show both theoretically and empirically that estimation variance tends to be higher in small subgroups \cite{agrawal2021debiasingclassifiersrealityvariance}.  Additionally, they prove results suggesting that partial debiasing results in both less variance and better fairness properties.

\section{Desiderata}
\label{sec:desiderata}

Although the problem of intersectionality is recognized in the literature, how difficult this problem is may not have been fully appreciated. At least some of the existing contributions violate minimal moral or methodological desiderata, as we shall see in Sections \ref{sec:minimal_justice}, \ref{sec:consistent_conceptualization}, and \ref{sec:incentive_compatibility} (and Appendix \ref{appendix:Kearns}).

A core tenet of building ethical algorithms is that machine-learned models need to be consistent with ``human values,'' which can be formulated as desiderata. We see the following desiderata for intersectional fairness metrics.

\subsection{Minimal Justice}\label{sec:minimal_justice}
A first desideratum we call ``minimal justice.'' The idea is, roughly, that a standard of fairness should not be lower for certain groups, such as those historically targeted for discrimination or facing structural injustice. Intuitively, minimal justice is a form of minority protection that says ``don't disadvantage the disadvantaged.'' 

This desideratum is a weak form of prioritarianism. Recent work in algorithmic fairness has identified a similar prioritarian idea in ``predictive justice'' \cite{lazar_site_nodate}. Whereas prioritarianism, a theory of distributive justice for well-being, demands that ``benefitting people matters more the worse off these people are'' \cite{parfit_equality_1997}, minimal justice requires only that those ``worse off'' should be given \emph{at least the same} weight in aggregating a fairness metric. The desideratum does \emph{not} require that greater weight be given to any group, and is hence met when a standard of fairness is identical for all groups. 

To illustrate the desideratum, consider an example. Notwithstanding its merits, the proposal of Kearns et al.\ \cite{Kearns_FairnessGerrymandering} may violate minimal justice. As noted above, the addition of $\alpha(G)$ in \eqref{eq:Kearns} relaxes the fairness constraint proportional to the size of a group. The smaller a group is (as a share of the data), the worse a model performance can be and still certify the model as fair. The fairness standard is hence lowered for small groups. On the assumption that these small groups include historically disadvantaged or oppressed groups, \eqref{eq:Kearns} violates minimal justice.

And drastically so: For a group $G'$ that is $c$ times smaller than group $G$ (i.e. $\frac{\alpha(G)}{\alpha(G')} = c)$, a model can be certified as ``fair'' if the disparity between the average performance and the performance for group $G'$ is as much as $c$-times worse than it is for group $G$. Furthermore, for some value of $\epsilon$ there are groups that are proportionally so small that there is no model performance poor enough to certify the model as unfair. For example, if $\epsilon = .01$, for a binary classifier, any group whose proportion of the total population is less than $\epsilon$ is protected by essentially no fairness constraint at all.\footnote{The average model performance $m(G)$ and $m(\cdot)$ is constrained to be less or equal to 1. But the maximum deviation of accuracy in the binary setting is 1. Thus, even if the model is entirely inaccurate for this population and perfectly accurate for the rest of the population, the constraint is still satisfied.} 

The ethical impact can be immense. A group might \emph{look} relatively small in the data but be, in fact, large in absolute numbers in the population. Indeed, disadvantaged groups tend to be under-represented in data \cite{lerman_big_2013, giest_for_2020}. Thus, the approach of Kearns et al.\ may lower the standard of fairness for precisely those groups that fairness is meant to protect.

\subsection{Consistent Conceptualization}\label{sec:consistent_conceptualization}
Any fairness metric operationalizes a certain idea, or concept, of fairness. A second desideratum is that fairness metrics should operationalize a concept of fairness consistently. 

This desideratum may resemble that of Minimal Justice. But whereas Minimal Justice is a moral desideratum, Consistent Conceptualization is a methodological one. Minimal Justice \emph{assumes} a standard of fairness as given (and requires that it not be lower for certain groups). Consistent Conceptualization ensures that this standard has minimal construct validity, i.e., that the formal operationalization of fairness represents the informal, intended conception of fairness (whichever that may be). The importance of construct validity for fairness is already established in the literature \cite{jacobs_measurement_2021}.

Typically, fairness metrics in algorithmic fairness operationalize the idea of \emph{equality}. This is particularly evident in \eqref{eq:typical} which, for each group $G$, restricts the absolute disparity of $m(G)$ from overall mean performance $m(\cdot)$. This is one---albeit a very simple---way of operationalizing inequality (for alternatives see \cite{sen_economic_1997}). Likewise, \eqref{eq:fairness_ratio} operationalizes fairness as equality \cite{9101635, AuditingAchievingIntersectional}.\footnote{Compared to \eqref{eq:typical}, \eqref{eq:fairness_ratio} aims for equality between groups (as opposed to minimizing disparity with $m(\cdot)$), and measures \emph{relative} disparity (a performance ratio instead of performance difference).} 

Moreover, \eqref{eq:typical} and the definition by Foulds et al.\ and Morina et al.\ operationalize equality \emph{consistently}. The fairness metrics apply an equality condition without bounds or exceptions.

Not so the proposal by Molina and Loiseau \cite{MolinaLoiseauBoundingApproximating}, which explicitly \emph{bounds} equality. Effectively, the fairness measure permits that some small number of people faces severe discrimination, as long as the likelihood of discrimination or their relative size as a share of the overall population is small.\footnote{ Molina and Loiseau \cite{MolinaLoiseauBoundingApproximating} write: ``It can be seen for some given $\epsilon$ as a statement on the expected size of the population that is not being discriminated too much against.''} This fairness metric thus fails the desideratum of operationalizing the concept of equality consistently. 

Typically fairness metrics, and all instances of \eqref{eq:typical}, operationalize fairness as equality. Alternatives, well-known from distributive justice, include \emph{prioritarianism}, stating that more of some good, such as model performance, should be given to those in greater need \cite{parfit_equality_1997}, and \emph{sufficientarianism}, requiring that everyone has \emph{enough} of some good (instead of the same) \cite{frankfurt_equality_1987, slote_beyond_1989}. 

\subsection{Incentive Compatibility}\label{sec:incentive_compatibility}

The final desideratum starts with the recognition that metrics specify incentives. Anyone who wants to increase their models' fairness may want to maximize a fairness metric. The final desideratum thus requires that a fairness metric not have ``perverse'' incentives of two kinds: discouraging data collection and allowing ``gaming.''

First, a fairness metric should not discourage data collection. Any fairness metric that indicates greater \emph{un}fairness only because further data are sampled from some group would fail to be incentive compatible. Likewise, inversely, any fairness metric would fail the desideratum that indicates greater \emph{fair}ness only because data based on group identity are dropped. 

The fairness metric \eqref{eq:Kearns}, of Kearns et al., likely violates this desideratum of incentive compatibility. This is because collecting more data on a minority population $G$ tightens the constraint by increasing $\alpha(G)$, thus making a certification of ``fairness'' at a given level of $\epsilon$ more difficult. Specifically, suppose that $m(G) = .15$ and $m(\cdot) = .85$. If $\alpha(G) = .01$, then the performance would be deemed ``fair'' for all  $\epsilon > 0.7 \times 0.01 = .007$. However, if we collect more data for group $G$ such that $\alpha(G) = .2$, then the model would be ``fair'' only for $\epsilon > 0.7 \times 0.2 = .014$. Unless the additional data results in material improvements to $m(G)$, for any $\epsilon$ such that $.007 < \epsilon < 0.014$, the fairness metric \eqref{eq:Kearns} would certify a given model as fair prior to further data collection, but as unfair afterwards. In short, under \eqref{eq:Kearns}, fairness for hard-to-predict groups could be attained simply by under-representing them in the training data. We see this effect in our empirical study, described in Appendix \ref{appendix:Kearns}.  We leave the details of this study to Appendix \ref{appendix:Kearns}, but the results show that the fairness metric suggested by Kearns et al.\ appears to indeed disincentivize additional data collection, violating \emph{Incentive Compatibility}.

This is a ``perverse'' effect because, in practice, additional data collection about a minority group will help improve the model performance for that group. In other words, the metric gives an incentive to do the opposite of what it is meant to achieve.\footnote{We do \emph{not} contend that more data should be collected. Privacy considerations are important. Our point is instead that maintaining the appearance of a good fairness metric is a bad reason to not collect more data.} 

Whether other metrics \cite{AuditingAchievingIntersectional, 9101635, MolinaLoiseauBoundingApproximating} violate this desideratum depends on whether the estimated performance disparity is greater than the true disparity (which further data would likely help approximate). Fairness metrics that operationalize fairness as \emph{equality} (e.g., as model performance disparity across groups), incentivize $m(G)$ to be nearly the same for all subgroups $G$. If the true model performance is nearly equal among groups, then these metrics incentive further data collection in order to have more accurate estimates of $m(G)$. 


Second, a fairness metric should not encourage knowingly erroneous predictions. But some metrics (e.g., statistical or demographic parity) have exactly this property: Even if the label that we want to predict is known (which it generally, of course, isn't), ``fairness'' as these metrics define it can be improved by erroneous predictions. This is an undesirable property of fairness metrics  \cite{dwork_fairness_2012}.


\section{Two Alternative Metrics}\label{sec:Our_models}

We now illustrate how these desiderata can be met. 
We propose two alternative models, which we call the ``optimist's'' and ``pessimist's model'' respectively. Both define the problem using hypothesis testing. The optimist has the null hypothesis that the model is fair, and we have to prove it is not (similar to ``innocent until proven guilty''); the pessimist inverts the ``burden of proof'' and has the null hypothesis that the model is unfair.\footnote{Throughout, we assume that for the metric $m(\cdot)$ larger values are better (think accuracy, not error rates). Specifically, and without loss of generality, we use accuracy as our sample metric.  This choice is for simplicity only; one could replace accuracy with any other metric for which higher values are preferred.  For the hypothesis tests we describe, we use a $z$-score of 1.64, which corresponds to a 95\% confidence interval for a one-sided hypothesis test. This is a conventional parameter choice and nothing in our argument depends on it.}


\subsection{Optimist's Model}\label{subsec:optimist}
We could formulate the problem of fairness for small groups as testing the joint hypothesis that 
\begin{align*}
    H_0 : m(G) > c \quad  \forall \, G\\
    H_1 : m(G) \leq c \quad \exists \, G
\end{align*}

Consider a group $G$ of size $n_G$. Suppose $m(G)$ is accuracy.  As a sample proportion, the standard error for our estimate of $m(G)$ is $\sqrt{\frac{m(G)(1-m(G))}{n_G}}$.  Then, we would reject the null if the upper end of its confidence interval is less than $c$, i.e., if $m(G) + 1.64\sqrt{\frac{m(G)(1-m(G))}{n_G}} < c$ (ignoring multiple testing).\footnote{This ignores multiple hypothesis testing, which we address in Appendix \ref{sec:limitations}.} Under this formulation, we reject $H_0$ if $m(G)$ is sufficiently less than $c$, where ``sufficiently less'' has to do with our statistical power to detect that it is less.  We would declare the model fair, if at given level $c$ we cannot statistically reject that the model performs at least $c$ well for all groups.  

A minority population which is sufficient in number would easily reject the null if  $m(G)$ is truly below $c$. Indeed, even with a population size of $n_G = 1000$, if $c = 0.7$, then a value of  $m(G) < 0.67$ would reject the hypothesis that the model is fair.

\subsection{Pessimist's Model}\label{subsec:pessimist}
Depending on a model's deployment context, the optimistic approach might be problematic.\footnote{Depending on the ethical risks involved in how a model is used, the more precautionary assumptions behind the pessimist's model might be more appropriate.} Consider instead the following pessimistic hypothesis test. 
\begin{align*}
    H_0 : m(G) < c \quad  \exists \, G\\
    H_1 : m(G) \geq c \quad  \forall \,  G
\end{align*}

We would declare the model fair, if at a given level $c$ we know with statistical certainty that the model performs at least $c$-well for all groups. In this case, (ignoring multiple testing again) we would require that $m(G) - 1.64\sqrt{\frac{m(G)(1-m(G))}{n_G}} > c$ for all $G$. 

\subsection{Fairness Metrics} 
The formulations can be extended from a hypothesis test to a fairness metric by finding the maximal $c$ for which the respective null hypothesis cannot be rejected (for the optimist) or can be rejected (for the pessimist). In the optimist's model, choose the maximal $c$ such that
\begin{equation}\label{eq:optimist}
    c \leq m(G) + 1.64\sqrt{\frac{m(G)(1-m(G))}{n_G}}
\end{equation}
for all relevant groups $G$. The fairness metric is the maximal $c$ such that we cannot reject the hypothesis that the model performs at least $c$-well for all groups.

This metric can be read as saying that a model is ``fair up to $c$.'' Intuitively, this means that, for all we know, the model performance $m(G)$ (say, accuracy) is likely as high as $c$ for each group.

On the pessimist's model, we instead choose the maximal $c$ such that
\begin{equation}\label{eq:pessimist}
    c \leq m(G) - 1.64\sqrt{\frac{m(G)(1-m(G))}{n_G}}
\end{equation}
for all relevant groups $G$. This fairness metric is the maximal $c$ such that we reject the hypothesis that the model is \emph{un}fair, that is, we reject that it does not perform at least $c$-well for each group. 

This metric can be read as saying that a model is ``unfair above $c$.'' The model likely performs at least $c$-well for each group; but for values above $c$, there likely is at least one group for which the model does not perform at least $c$-well---and we hence can't rule out that the model is unfair.

In summary, the fairness metrics are defined as bounds of the interval 
\begin{equation*}
    \left( m(G) - 1.64\sqrt{\frac{m(G)(1-m(G))}{n_G}}, m(G) + 1.64\sqrt{\frac{m(G)(1-m(G))}{n_G}}\right)
\end{equation*}
This interval, of course, now has two interpretations. For one, it is the 90\% confidence interval for the value of $m(G)$ for each $G$. Moreover, across all groups, it is also the interval in which we cannot reject the hypothesis that the model is unfair, nor can we reject the hypothesis that the model is fair.\footnote{The reader may go to Appendix \ref{sec:impact_m_n} for an exploration on the impacts of changing $m$ and $n$ on these models.}

\subsection{Discussion: Desiderata} \label{section:optimist_vs_cynic}

Both metrics satisfy \emph{Minimal Justice}. The bound $c$ encodes a standard of fairness that is identical for all groups. Moreover, the relative size of groups doesn't matter. Whether a null hypothesis can be rejected changes with the absolute size of the group $n_G$ (rather than the proportion $\frac{n_G}{n}$). 

On the optimist's metric, for a small group, the difference between the actual (lower) model performance and the level up to which a model can be certified as fair might be large. But both of our metrics base their certification of ``fairness up to $c$'' on an aggregation that gives all groups the same weight. In fact, the pessimist's metric can be called ``epistemically risk averse'' insofar as it picks the \emph{highest lower} bound out of all groups' confidence intervals (and hence is similar to the maximin decision rule).

On \emph{Consistent Conceptualiztion}, both of our metrics conceptualize fairness as sufficiency. They understand fairness not as a matter of whether everyone has the same (as equality does), but whether everyone has \emph{enough} \cite{frankfurt_equality_1987,slote_beyond_1989}. This idea is operationalized in \eqref{eq:optimist} and \eqref{eq:pessimist} in a transparent and natural way: with an inequality. Moreover, the threshold $c$, what counts as ``enough,'' is determined absolutely in the terms of model performance measure, and not depending on, e.g., how well the model performs on other groups. Thus, both of our metrics operationalize sufficiency consistently across all groups.

For \emph{Incentive Compatibility} the picture is mixed:
Both of our metrics discourage gaming (and thus satisfy Incentive Compatibility in this respect). This is because both fairness metrics determine (un)fairness as the highest (or lowest) expectable model performance across all groups. As such, improving model performance will never increase unfairness; and decreasing model performance will never increase fairness. In fact, decreasing model performance may lead to a decrease in fairness. 
It appears that operationalizing the idea of fairness as sufficiency is what makes our fairness metrics less susceptible to gaming---in particular, that the minimum level of model performance is defined in absolute terms and equally enforced for all groups.  

But one of our metrics, namely the optimist's, may discourage further data collection (and thus violate Incentive Compatibility in its first respect). Because the optimist's model starts with the null hypothesis that a model is fair at a given $c$, gathering more data can make things ``worse''; that is, with more data, we might come to reject the optimistic null hypothesis of fairness at a given $c$.  
A model might perform very poorly for certain groups, but we cannot reject the null hypothesis that the model is fair up to $c$, thanks to sparse data---and the metric thus results in an incentive to not sample more data but to instead ``look the other way.''

\section{Fairness Datasets Analysis}\label{sec:analysis}

We evaluate empirically whether our metrics meet the desideratum of incentive compatibility. The question is: Do our metrics incentivize or disincentivize additional data collection?

To answer this question, we ``simulate'' additional data collection by experiment. We train models on increasingly larger subsamples of benchmark datasets and observe how metrics behave as the size of the training data increases. The behavior that we want to see is that the fairness metrics increase with the size of the training data sampled from the dataset. If, instead, a fairness metric \emph{decreased} as greater shares of the dataset are sampled, the metric would \emph{disincentivize} further data collection.  

We seek to observe our metrics' behavior across the largest feasible range of benchmarks. To achieve this, we use lale, a Python library created by IBM \cite{lale_library}. Lale allows for the creation of consistent automated machine learning models across 20 well-known ``fairness datasets'' that can easily be fetched, modeled, and evaluated \cite{lale_fairness_datasets}. These datasets are all tabular with a categorical target variable. They each come with ``fairness metadata,'' which includes protected attributes, along with ranges/values of those attributes that correspond to the privileged group.\footnote{While there have been critiques of the usage of some of these datasets \cite{ding2021retiring} \cite{compaslicated}, they are still appropriate for the purpose of testing whether our proposals incentivize or disincentivize the collection of additional data.} Details on the methods of our analysis are in Appendix \ref{appendix:methods}. Here we only discuss the main result on testing whether our metrics incentivize against data collection. 

For each of the datasets, we observe model performance $m(G)$, as well the optimist's $c_1^g$ and the pessimist's fairness metric $c_2^g$ respectively. For ease of interpretation we use accuracy as model performance; neither our results nor their interpretation depend on this.

\begin{figure}[ht]
\centering
\includegraphics[width=\linewidth]{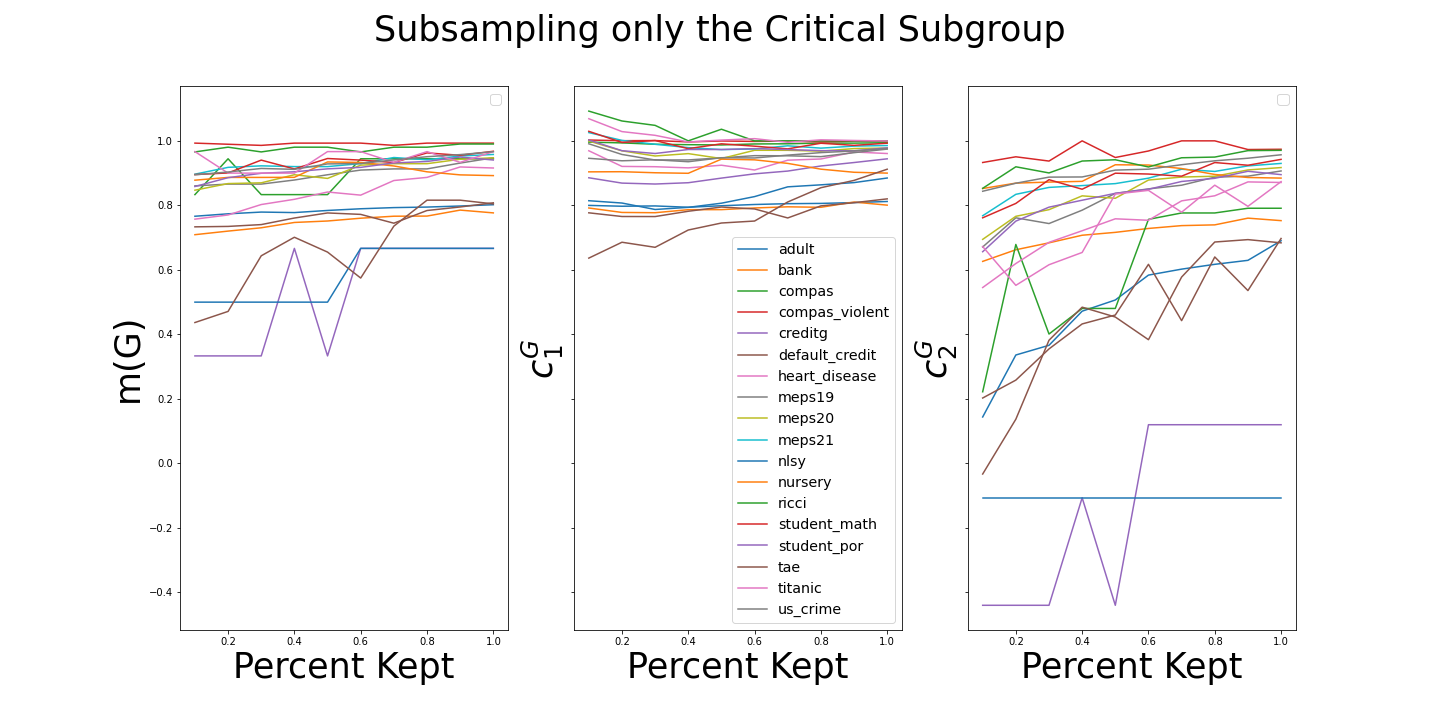}
\caption{Plots of accuracy $m(G)$, optimist's metric $c_1^g$, and pessimist's metric $c_2^g$ of critical subgroups $G$ for each dataset. The $x$-axis corresponds to the percentage of the critical subgroup that is kept. Legend lists the dataset name.}
\label{graphs:subsample_borderline}
\end{figure}

We ran two versions of this experiment. In one version, we subsample the entire dataset of each benchmark; whereas in another, we subsample only on the \emph{critical subgroup}, which is the group that is right on the $c$ threshold. The first version simulates additional data collection for \emph{all} groups, whereas the latter for those groups that ``drag down'' the fairness metric. Here we concentrate on results from subsampling on the critical subgroup only, shown in Figure \ref{graphs:subsample_borderline}.\footnote{Some of the plots are disconnected.  This is because sometimes the subsampling of the dataset did not include any members of the critical subgroup; in those cases, the model could not predict for that subgroup, so no accuracy measurement could be taken.  The most erratic curves (curves of $m(G)$ and $c_2$ for the {\tt creditg} and {\tt nlsy} datasets) correspond to either a subgroup of size 1 or 2.} Full results for both versions are in Appendix \ref{appendix:analysis}.

The thing to note here is that there is a trend upwards in each of these plots. Most notably, the middle plot, on the optimist's metric $c_1^g$ shows this upward trend.\footnote{Some datapoints ``overshoot'' on the $y$-axis with values $>1$, suggesting a negative trend, e.g., for the {\tt compas} dataset (green line). But this behavior is an artifact of the standard way of calculating the confidence interval.} This suggests that our optimist's metric---at least for the datasets tested---does \emph{not} pose perverse incentives. The further we go on the x-axis (representing more data being ``collected''), the model performance as well as the fairness metrics tend to improve. 

Consider for example the behavior of the optimist's metric for the model trained on increasing amounts of data from the {\tt tae} dataset (brown line that ``starts'' lowest in middle figure). Although the metric does not strictly increase as the training is based on greater data (the metric decreases slightly from 20\% to 30\% of data used), it shows a very strong upward trend.

\section{Conclusion}

Although the general idea of intersectionality seems easy to state, putting intersectionality to work in quantitative social science is, generally, far from straight-forward \cite{bright_causally_2016}. Likewise, intersectionality presents a problem for algorithmic fairness: Intersectionality requires estimating statistical properties across subgroups that are increasingly small, which gives rise to statistical as well as moral-methodological challenges.

Statistically, small groups are a challenge for estimation.  As statistical uncertainty increases (due to more and smaller groups), the point estimates of model performance for these groups become meaningless. Any approach of intersectional fairness needs to account for statistical uncertainty. But some existing metrics do not seem to fully appreciate the moral-methodological challenges that underlie this problem and ``lower the fairness bar'' for smaller groups, i.e., the metrics violate desiderata such as Minimal Justice or Consistent Conceptualization.


With this paper, we elucidate this intersectionality problem for algorithmic fairness: We develop minimal desiderata to clarify the moral-methodological challenges underlying this problem; we argue that some existing fairness metrics fail these desiderata, but illustrate that the desiderata can be met. We propose fairness metrics that rely on hypothesis testing (instead of performance point estimates) and that understand fairness as sufficiency (instead of equality). On these proposed metrics, fairness is understood as a certain minimum level of expected model performance that is, for all we know, likely enjoyed by all groups. We empirically evaluate the metrics against the proposed desiderata, including on 18 datasets that are widely used for fairness benchmarks.

In light of their technical and normative-theoretical limitations, the metrics we propose should be seen as illustrations. Technically, the simple hypothesis testing needs to be extended to multiple hypothesis testing to allow for interdependent subgroup memberships (see Appendix \ref{sec:limitations}). Normative-theoretically, the desiderata that we develop are not exhaustive and they do not uniquely characterize the metrics we propose.

Nevertheless, overall, our findings extend the list of problems that statistical uncertainty raises for algorithmic fairness. Previous work observed that fairness metrics are biased: They ``fail to account for statistical uncertainty\,\dots exaggerating the extent of performance disparities'' between groups where such disparities exist and indicating disparities ``in cases where model performance is\,\dots identical across groups'' \cite{lum_-biasing_2022}. Our present findings add that with increasing statistical uncertainty fairness metrics risk becoming either nonsensical (if they aggregate point estimates) or morally inadequate (if they ``lower the fairness bar'' to enable statistical estimation).

However, we also offer ways of advancing the literature on algorithmic fairness: with desiderata that clarify the challenges at hand and guide the search for solutions, and with fairness metrics that suggest novel avenues for defining such metrics based on hypothesis testing and fairness as sufficiency.

\section*{Acknowledgements}

This material is based upon work supported by the National Science Foundation under Grant No. DMS-1928930 and by the Alfred P. Sloan Foundation under grant G-2021-16778, while Ellen Veomett was in residence at the Simons Laufer Mathematical Sciences Institute (formerly MSRI) in Berkeley, California, during the Fall 2023 semester.

\printbibliography

\newpage 

\appendix

\section{Discussion: Impact of \texorpdfstring{$n$}{n} and \texorpdfstring{$m$}{m} on Each Model}\label{sec:impact_m_n}

\begin{figure}[ht]
\begin{subfigure}[t]{0.32\linewidth}
    \includegraphics[width=\textwidth]{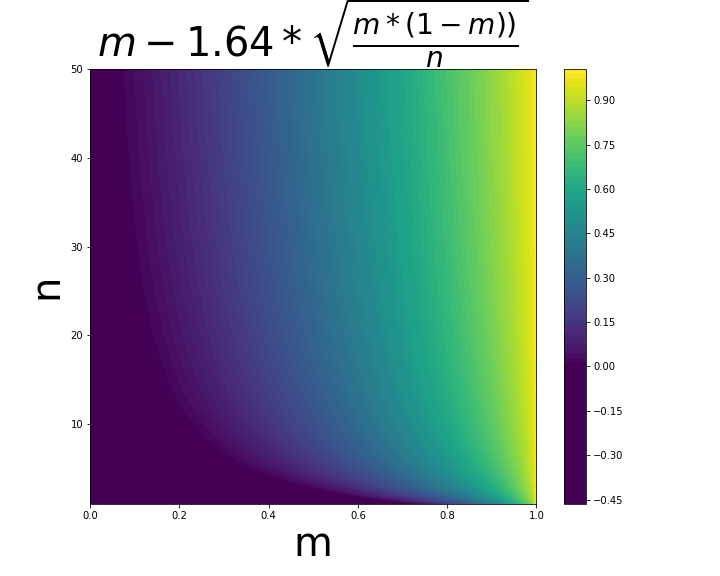}
    \caption{Values of $m-1.64\sqrt{\frac{m(1-m)}{n}}$}
    \label{fig:2dlower}
    \end{subfigure}
    \begin{subfigure}[t]{0.32\linewidth}
    \includegraphics[width=\textwidth]{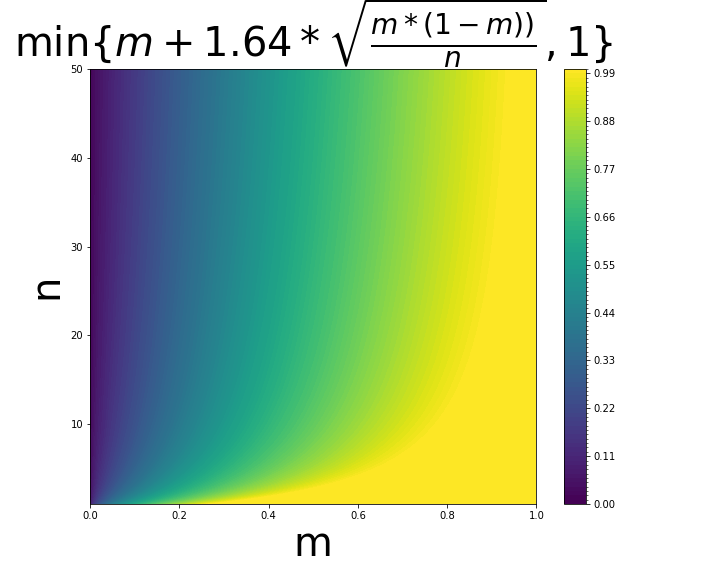}
    \caption{Values of \\ $\min\{m+1.64\sqrt{\frac{m(1-m)}{n}}, 1\}$}
    \label{fig:2dupper_with_min}
    \end{subfigure}
    \begin{subfigure}[t]{0.32\linewidth}
    \includegraphics[width=\textwidth]{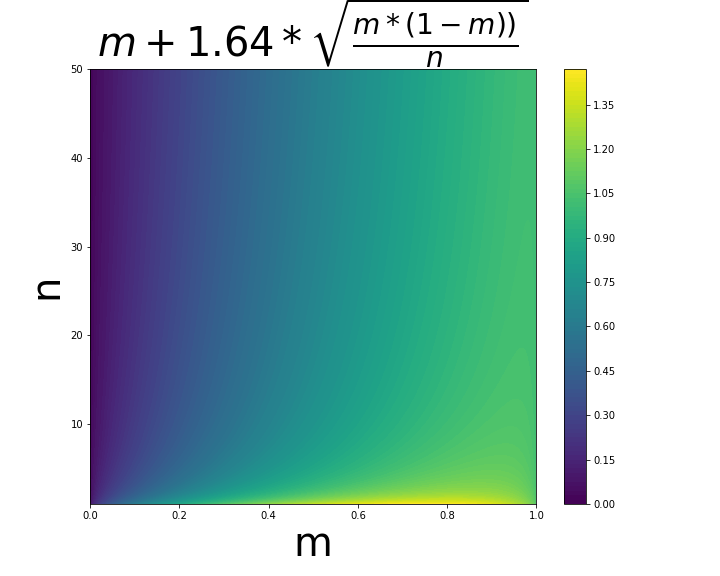}
    \caption{Values of $m+1.64\sqrt{\frac{m(1-m)}{n}}$}
    \label{fig:2dupper}
    \end{subfigure}
    \caption{Showing the relationship between $m$ (metric), $n$ (number in subgroup), and $c$ (edge of confidence interval).  Hues in \ref{fig:2dlower} shows the values of $c$ in the pessimist's model.  Hues in \ref{fig:2dupper_with_min} shows the values of $c$ in the optimist's model, but with an upper limit of 1 (since no proportion can be larger than 1).  Hues in \ref{fig:2dupper} shows the values of $c$ in the optimist's model, without limiting the value at 1 (so that we can see more easily where it is very difficult to reject the optimist's hypothesis that the model is fair).}
    \label{fig:2dplots}
\end{figure}

To give the reader a feel for the mathematical impact of the choice between these two models, we share some hopefully informative plots in Figure \ref{fig:2dplots}.

The horizontal axis of each of these plots is $m$, the metric value, which is assumed to be a proportion for which higher values are preferred (such as accuracy).  The vertical axis is $n$, the size of the subgroup.  The hue at $(m,n)$ in Figure \ref{fig:2dlower} is the corresponding value of $m-1.64\sqrt{\frac{m(1-m)}{n}}$.  Here we can visually see that, for a fixed metric value $m$, the subgroup size must be reasonably large in order to reject the hypothesis that the model is ``unfair above $c$'' for $c$ near $m$.

Similarly, the hue at $(m,n)$ in Figure \ref{fig:2dupper_with_min} is the corresponding value of $m+1.64\sqrt{\frac{m(1-m)}{n}}$, capped at a value of 1 (since no proportion can be larger than 1).  Here we can visually see that, for a fixed metric value $m$, the subgroup size must be reasonably large in order to reject the hypothesis that the model is ``fair up to $c$'' for $c$ near $m$.  To further understand the impact of small groups in this optimist's model, we include Figure \ref{fig:2dupper}.  In Figure \ref{fig:2dupper}, the hue simply gives the value of $m+1.64\sqrt{\frac{m(1-m)}{n}}$, even if it is larger than 1.  This plot further highlights the fact that, in the optimist's model, it is very difficult to reject the hypothesis that the model is perfectly fair for very small subgroups.

\section{Limitations}\label{sec:limitations}
We note that the issue of multiple hypothesis testing is one which we do not address in depth. If membership in the different groups in question is independent, one can use the Bonferroni correction to address the multiple hypothesis tests.  Under this strict type of multiple hypothesis testing, the p-values that are calculated are using significance level $\frac{\alpha}{n}$, where $n$ is the number of hypotheses that we are testing.  This correction guarantees that the probability that we reject \emph{one or more} null hypotheses is no more than $\alpha$. Considering overlapping subgroups (such as considering fairness both for Black Women and for Latina Women) requires more care, and we do not delve into the issue of overlapping subgroups here. We thus, effectively, assume---counterfactually---that each person is a member of exactly one group. For the purposes of our empirical study (below), we fix the number of protected attributes to be as large as possible, as described in Section \ref{sec:subgroupd_id}. 

\section{Methods}\label{appendix:methods}
We here provide further details on our empirical methods.

For starters, we choose the lale library and its accompanying datasets for two reasons:
\begin{enumerate}
    \item The number of ``fairness datasets'' in the lale library is larger than any other conglomeration of fairness datasets that we are aware of.
    \item Because the lale library has built-in models, we can apply a consistent type of model to each dataset, so that our experiments are not muddied by differing model constructions.
\end{enumerate}

\subsection{Subgroup Identification}\label{sec:subgroupd_id}

The models we created use a forest of boosted trees from the XGBoost library; the functions to easily create these models are also part of the lale library. 
We created three models using the lale pipeline, using 3-fold cross-validation.  The three models can be accessed to evaluate their accuracy on various subgroups.  However, since lale requires sklearn version 1.2, we do not have access to the train/test indices of each of the models.  Thus, to evaluate the accuracy on group $G$, we do so on all of the members of $G$ in the dataset.\footnote{Ideally, we would like to evaluate only on the members of $G$ in the test set for that fold.  However, our goal here is to assess our two proposed ideas to address small-sized subgroups, not to assess true model accuracy.  Averaging the subgroup accuracy across the three folds provides appropriate information to do that.  Thus, for this analysis, we set $m(G)$ to be the average accuracy of the three models for subgroup $G$.}

The set of subgroups $G$ on which we calculated the model accuracy come in part from the fairness data that lale provides, and also from attributes that are well-understood to be sensitive.  Specifically, all of the attributes that the lale library lists as ``protected'' are included in our master list of protected attributes.  If the rows in the dataset correspond to individuals, and any of $\{\text{age}, \text{sex}, \text{race}\}$ were not in lale's list of protected attributes, we added them to the master list.  From this master list, we created \emph{all} subgroups using \emph{all} categories in the master list.  For example, if a dataset had race, sex, and age category,  we included in $G$ each triple $(r, s, a)$, where $r$ was a race in that dataset's race column, $s$ was a sex in that dataset's sex column, and $a$ was an age category for that dataset. 

\subsection{Data Pre-processing}
For each of the 20 fairness datasets, we used the built-in lale data pre-processing with small adjustments.

We used the simple methods for imputing missing data which are provided with the sample notebook at \cite{lale_fairness_notebook}. 

In order to use XGBoost, we needed to change some of the predicted categories to integer type.  

In order to make the results more understandable, we re-named some of the categories (for example, changing the `sex' categories from 0/1 to male/female). 

The ``race'' categories in the {\tt nlsy} dataset were atypical, including both categories such as `GERMAN' and `BLACK.'  We did not attempt to clean that data but left the categories as given.

We created groupings by age for those datasets that don't already come with age groupings (see Appendix \ref{sec:age-group-heuristic}). 

\subsection{Age Grouping}\label{sec:age-group-heuristic}
For the age attribute, some of the datasets already come with age groupings.  In those cases, we directly used those groupings as the age categories.  For the datasets where age was a strictly numerical attribute, we used the following heuristic to create categories:
\begin{itemize}
\item If age was already listed by lale as a protected attribute, we used the ranges provided by lale (for priviledged/unpriviledged groups) to create the categories.
\item If age was not already listed as a protected attribute:
\begin{itemize}
\item We grouped by decade in all datasets where this produced at least 5 people of each decade.
\item The {\tt law\_school} dataset had fewer than 5 members of the [0,9] decade, and fewer than 5 members of the [10, 19] decade, so those were grouped into a 0-19 group
\end{itemize}
\end{itemize}

After this initial analysis, we tossed out two of the datasets: {\tt law\_school} and {\tt speeddating}.  The standard lale models created by XGBoost were 100\% accurate on those models, and thus did not provide interesting analysis for us.\footnote{We suspect that these datasets might be included in lale's list because they have low scores on other fairness metrics, such as the ``symmetric class imbalance'' metric in the sample notebook at \cite{lale_fairness_notebook}, or because the model must use protected attributes in order to be accurate.}

\subsection{Analysis}
\label{appendix:analysis}

For each such subgroup $G \in \mathcal{G}$, we calculated $m(G)$: the average accuracy of the three models on that subgroup.  We then calculate the $c$ values associated with each of those subgroups; indexed by $c_1$ for the optimist's and $c_2$ for the pessimist's metric.  Specifically, for group $G$ we calculate
\begin{equation*}
    c_1^G =m(G) + 1.64\sqrt{\frac{m(G)(1-m(G))}{n_G}}
\end{equation*}
from the optimist's model and
\begin{equation*}
    c_2^G =m(G) - 1.64\sqrt{\frac{m(G)(1-m(G))}{n_G}}
\end{equation*}
from the pessimist's model. 

Once these are calculated for all subgroups, we calculate
\begin{align*}
    acc_{min} &= \min\{m(G): G \in \mathcal{G}\} \\
    c_1 &= \min\{c_1^G: G \in \mathcal{G}\} \\
    c_2&= \min\{c_2^G: G \in \mathcal{G}\} \\
\end{align*}
We also find their corresponding subgroups:
\begin{align*}
    G_{min\_acc} &= \text{argmin}\{m(G): G \in \mathcal{G}\} \\
    G_1 &= \text{argmin}\{c_1^G: G \in \mathcal{G}\} \\
    G_2 &= \text{argmin}\{c_2^G: G \in \mathcal{G}\} \\
\end{align*}
The group $G_{min\_acc}$ is the group with minimum estimated accuracy, while group $G_1$ ($G_2$) is on the cusp of rejecting the hypothesis that the model is fair (not being able to reject the hypothesis that the model is unfair) up to accuracy $c_1$ ($c_2$).  Thus, we call groups $G_{min\_acc}, G_1$, and $G_2$ the \emph{critical subgroups} for a dataset.  For some datasets, there are three distinct critical subgroups, while for other datasets, some of the critical subgroups are the same; see Tables \ref{table:min_subgroups}, \ref{table:acc_details}, \ref{table:c1_details}, and \ref{table:c2_details} in Appendix \ref{appendix:tables} for details.

Once we had the (up to) three critical subgroups of each dataset, we did two additional analyses.

\subsubsection{Subsample Just the Critical Group}\label{sec:subsample_crit} Suppose $G$ is a critical subgroup of a dataset.  We then created 10 models (each a set of three 3-fold cross-validated models), where we include 10\%, 20\%, $\dots$, 100\% of the subgroup in the dataset used to create the model.  We then evaluated that group's critical value (whether it be $m(G)$, $c_1^G$, or $c_2^G$) on each of those 10 models, to see how those values change.  The intention here is to mimic increasing samples from just the critical group, and how that additional data collection impacts the fairness evaluation of the model.  These results of this analysis were in Figure \ref{graphs:subsample_borderline}.  

\subsubsection{Subsample the Entire Dataset}\label{sec:subsample_all}   
\begin{figure}[ht]
\centering
\includegraphics[width=\linewidth]{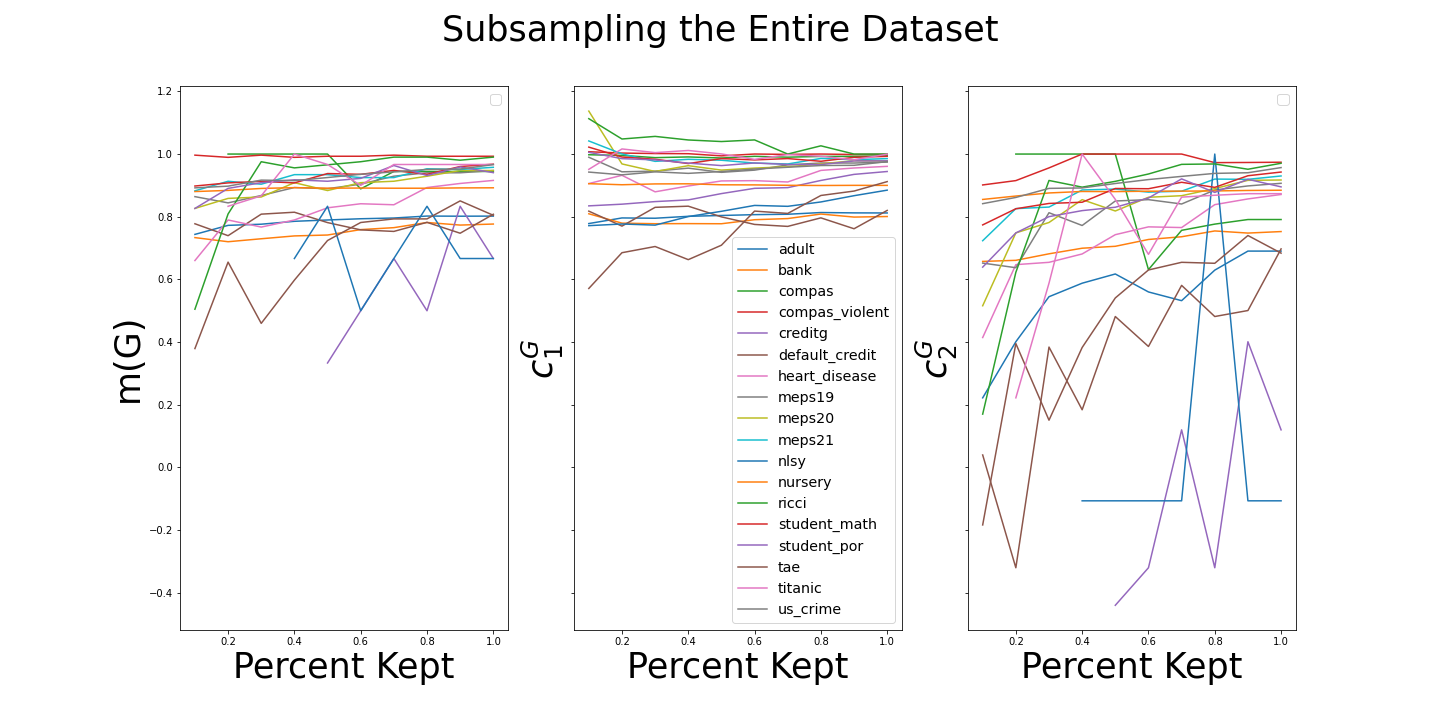}
\caption{Plots of $m(G), c_1^G$, and $c_2^G$ of critical subgroups $G$ for each dataset.  Here we subsampled the entire dataset, and the $x$-axis corresponds to the percentage of the entire dataset that is kept.  Legend lists the dataset name.}
\label{graphs:subsample_all}
\end{figure}

Suppose $G$ is a critical subgroup of a dataset.  We also created 10 models (each a set of three 3-fold cross-validated models), where we included 10\%, 20\%, $\dots$, 100\% of the entire dataset to create the model.  We then evaluated that group's critical value (whether it be $m(G)$, $c_1^G$, or $c_2^G$) on each of those 10 models, to see how those values change.  The intention here is to mimic increasing sampling overall, and how that additional data collection impacts the fairness evaluation of the model.  We note that, for the {\tt nursery} dataset, one of the predicted categories (recommend) had only two data points with that category.  In order for XGBoost to successfully create a model, we needed to add back both of those two data points into each subsample (if they had been removed in that random subsample).  The results of this analysis are in Figure \ref{graphs:subsample_all}.

\clearpage 

\section{Tables}\label{appendix:tables}

\begin{table}[ht]
\begin{tabular}{lp{3cm}p{3cm}p{3cm}} \hline 
Dataset         & Min $m(G)$ Subgroup                                   & Min $c_1$ Subgroup               & Min $c_2$ Subgroup                         \\ \hline  \hline 
adult           & (White, Male, 50's)                      & (White, Male, 50's)    & (Other, Male, 60's)              \\ \hline 
bank            & \textless{}=24                                 & \textless{}=24               & \textless{}=24                         \\ \hline 
compas          & (Male, Native American, 25 - 45)         & (Male, African-American, 25 - 45) & (Male, Native American, 25 - 45) \\ \hline 
compas\_violent & (Female, African-American, $<$25) & (Male, African-American, 25-45)                         & (Male, Other, $>$45)   \\ \hline 
creditg         & (male div/sep, male, \textless{}=25)             & (male single, male, $>$25)               & (male div/sep, male, \textless{}=25)     \\ \hline 
default\_credit & (male, 40's)                               & (male, 40's)                         & (female, 70's)                     \\ \hline 
heart\_disease  & (female, \textgreater{}54)                 & (female, \textgreater{}54)                       & (female, \textgreater{}54)         \\ \hline 
meps19          & (White, 80's, female)                           & (White, 80's, female)                  & (Non-White, 80's, male)               \\ \hline 
meps20          & (Non-White, 80's, female)                       & (Non-White, 80's, female)                         & (Non-White, 80's, female)               \\ \hline 
meps21          & (Non-White, 80's, female)                       & (Non-White, 80's, female)                         & (Non-White, 80's, female)               \\ \hline 
nlsy            & (Female, \textless{}18, GREEK)                  & (Female, $>$=18, GERMAN)         & (Female, \textless{}18, HAWAIIAN)          \\ \hline 
nursery         & great\_pret                                    & great\_pret                  & great\_pret                            \\ \hline 
ricci           & W                                              & B                            & W                                      \\ \hline 
student\_math   & (M, \textless{}18)                         & (M, \textless{}18)                & (M, \textless{}18)                 \\ \hline 
student\_por    & (M, \textgreater{}=18)                     & (M, \textless{}18)                           & (M, \textgreater{}=18)             \\ \hline 
tae             & 1.0                                            & 2.0                          & 1.0                                    \\ \hline 
titanic         & (female, 60's)                             & (female, 30's)                       & (female, 60's)                     \\ \hline 
us\_crime       & TRUE                                           & TRUE                         & TRUE    \\ \hline                               
\end{tabular}
\caption{Subgroups with minimum $m(G)$, $c_1$, or $c_2$ for each dataset.}
\label{table:min_subgroups}
\end{table}

\begin{table}[ht]
\begin{tabular}{lp{3cm}p{3cm}p{1cm}l}\hline
Dataset         & Subgroup                                   & Subgroup Category                                           & n & $m(G)$   \\ \hline \hline
adult           & (White, Male, 50's)                      & {[}race, sex, age\_cat{]}                            & 4256              & 0.8020050125313280 \\ \hline
bank            & \textless{}=24                                 & age\_cat                                                   & 809               & 0.7766790276060980 \\ \hline
compas          & (Male, Native American, 25 - 45)         & {[}sex, race, age\_cat{]}                            & 6                 & 0.9444444444444450 \\ \hline
compas\_violent & (Female, African-American, $<$25) & {[}sex, race, age\_cat{]}                            & 95                & 0.9929824561403510 \\ \hline
creditg         & (male div/sep, male, \textless{}=25)             & {[}personal\_status, sex, age\_cat{]}                       & 2                 & 0.6666666666666670 \\ \hline
default\_credit & (male, 40's)                               & {[}sex, age\_cat{]}                                    & 2771              & 0.8078912546613740 \\ \hline
heart\_disease  & (female, \textgreater{}54)                 & {[}sex, age\_cat{]}                                    & 103               & 0.9158576051779940 \\ \hline
meps19          & (White, 80's, female)                           & {[}RACE, age\_cat, SEX{]}                            & 184               & 0.947463768115942  \\  \hline
meps20          & (Non-White, 80's, female)                       & {[}RACE, age\_cat, SEX{]}                            & 146               & 0.9474885844748860 \\ \hline
meps21          & (Non-White, 80's, female)                       & {[}RACE, age\_cat, SEX{]}                            & 142               & 0.9577464788732400 \\ \hline
nlsy            & (Female, \textless{}18, GREEK)                  & {[}gender, age\_cat, race{]}                                   & 2                 & 0.666666666666667 \\ \hline
nursery         & great\_pret                                    & parents                                                    & 4320              & 0.8922839506172840 \\ \hline
ricci           & W                                              & race                                                       & 68                & 0.9901960784313730 \\ \hline
student\_math   & (M, \textless{}18)                         & {[}sex, age\_cat{]}                                    & 134               & 0.9676616915422890 \\ \hline
student\_por    & (M, \textgreater{}=18)                     & {[}sex, age\_cat{]}                                    & 73                & 0.9406392694063930 \\ \hline
tae             & 1.0                                            & whether\_of\_not \_the\_ta\_is\_a\_native \_english\_speaker & 29                & 0.8045977011494250 \\ \hline
titanic         & (female, 60's)                             & {[}sex, age\_cat{]}                                    & 10                & 0.9666666666666670 \\ \hline
us\_crime       & TRUE                                           & blackgt6pct                                                & 970               & 0.9663230240549830 \\ \hline
\end{tabular}
\caption{Subgroups with minimum accuracy value $m(G)$}
\label{table:acc_details}
\end{table}

\begin{table}[ht]
\begin{tabular}{lp{3cm}p{3cm}p{1cm}l}\hline 
Dataset         & Subgroup               & Subgroup Category                                         & n & $c_1$ \\ \hline\hline
adult           & (White, Male, 50's)    & {[}race, sex, age\_cat{]}                            & 4256                 & 0.8120224969943970  \\ \hline
bank            & \textless{}=24               & age\_cat                                                   & 809                  & 0.8006925092362380  \\ \hline
compas          & (Male, African-American, 25 - 45) & {[}sex, race, age\_cat{]}                                        & 1563                & 0.994278290695671    \\ \hline
compas\_violent & (Male, African-American, 25 - 45)                        & {[}sex, race, age\_cat{]}                                                        & 932                 & 0.999219907516058  \\ \hline
creditg         & (male single, male, $>$25)               & [personal\_status, sex, age\_cat]                                                   & 492                  & 0.94429531762294  \\ \hline
default\_credit & (male, 40's)                         & (sex, age\_cat)                                                   & 2771                 & 0.820164957561691  \\ \hline
heart\_disease  & (female, $>$54)                       & (sex, age\_cat)                                                        & 103                  & 0.96071630150011  \\ \hline
meps19          & (White, 80's, female)                  & {[}RACE, age\_cat, SEX{]}                                    & 184                  & 0.974437789794083  \\ \hline
meps20          & (Non-White, 80's, female)                         & {[}RACE, age\_cat, SEX{]}                                                    & 146                  & 0.977763384001414  \\ \hline
meps21          & (Non-White, 80's, female)                         & {[}RACE, age\_cat, SEX{]}                                                    & 142                  & 0.985432236390149  \\ \hline
nlsy            & (Female, $>$=18, GERMAN)         & {[}gender, age\_cat, race{]}                                     & 179                 & 0.88480628727837  \\ \hline
nursery         & great\_pret                  & parents                                                    & 4320                 & 0.9000195456124770  \\ \hline
ricci           & B                            & race                                                       & 27                   & 1.0                 \\ \hline
student\_math   & (M, \textless{}18)                & (sex, age\_cat)                                                   & 134                  & 0.992723469193729  \\ \hline
student\_por    & (M, \textless{}18)                            & (sex, age\_cat)                                                        & 193                  & 0.978258629541195  \\ \hline
tae             & 2.0                          & whether\_of\_not \_the\_ta\_is\_a\_native \_english\_speaker & 122                  & 0.912074688695555   \\ \hline
titanic         & (female, 30's)                       & (sex, age\_cat)                                                        & 86                  & 0.99964644295967    \\ \hline
us\_crime       & TRUE                         & blackgt6pct                                                & 970                  & 0.975822194666121  \\ \hline
\end{tabular}
\caption{Subgroups with minimum $c_1$ value}
\label{table:c1_details}
\end{table}

\begin{table}[ht]
\begin{tabular}{lp{3cm}p{3cm}p{1cm}l} \hline 
Dataset         & Subgroup                         & Subgroup Category                                         & n & $c_2$  \\ \hline \hline
adult           & (Other, Male, 60's)              & {[}race, sex, age\_cat{]}                            & 10                   & 0.6903719639038720   \\ \hline
bank            & \textless{}=24                         & age\_cat                                                   & 809                  & 0.7526655459759590   \\ \hline
compas          & (Male, Native American, 25 - 45) & {[}sex, race, age\_cat{]}                            & 6                    & 0.7910815916767240   \\ \hline
compas\_violent & (Male, Other, Greater than 45)   & {[}sex, race, age\_cat{]}                            & 49                   & 0.9739395574376140   \\ \hline
creditg         & (male div/sep, \textless{}=25)     & {[}personal\_status, age\_cat{]}                       & 2                    & 0.12000000000000000  \\ \hline
default\_credit & (female, 70's)                     & {[}sex, age\_cat{]}                                    & 12                   & 0.6973855176357850   \\ \hline
heart\_disease  & (female, \textgreater{}54)         & {[}sex, age\_cat{]}                                    & 103                  & 0.8709989088558780   \\ \hline
meps19          & (Non-White, 80's, male)               & {[}RACE, age\_cat, SEX{]}                            & 67                   & 0.9066848240754210   \\ \hline
meps20          & (Non-White, 80's, female)               & {[}RACE, age\_cat, SEX{]}                            & 146                  & 0.9172137849483570   \\ \hline
meps21          & (Non-White, 80's, female)               & {[}RACE, age\_cat, SEX{]}                            & 142                  & 0.9300607213563300   \\ \hline
nlsy            & (Female, \textless{}18, HAWAIIAN)          & {[}sex, age\_cat, race{]}                                   & 1                    & -0.10643674743062500 \\ \hline
nursery         & great\_pret                            & parents                                                    & 4320                 & 0.8845483556220900   \\ \hline
ricci           & W                                      & race                                                       & 68                   & 0.9706008691220500   \\ \hline
student\_math   & (M, \textless{}18)                 & {[}sex, age\_cat{]}                                    & 134                  & 0.9425999138908480   \\ \hline
student\_por    & (M, \textgreater{}=18)             & {[}sex, age\_cat{]}                                    & 73                   & 0.8952823458833590   \\ \hline
tae             & 1.0                                    & whether\_of\_not \_the\_ta\_is\_a\_native \_english\_speaker & 29                   & 0.6838443819651760   \\
titanic         & (female, 60's)                     & {[}sex, age\_cat{]}                                    & 10                   & 0.8735726878662690   \\ \hline
us\_crime       & TRUE                                   & blackgt6pct                                                & 970                  & 0.9568238534438450  \\ \hline
\end{tabular}
\caption{Subgroups with minimum $c_2$ value}
\label{table:c2_details}
\end{table}

\clearpage 

\section{Analysis of Metric from Kearns et al.}\label{appendix:Kearns}

As noted in Section \ref{sec:incentive_compatibility}, we hypothesize that the fairness metric outlined by Kearns et al.\ \cite{Kearns_FairnessGerrymandering} violates \emph{Incentive Compatibility}. The fairness metric likely ``looks worse'' as additional data is gathered about a small subgroup (i.e., a group whose size in proportion to the entire dataset is small). The fairness metric includes a factor which is the proportion of the subgroup within the dataset. Thus, as additional data is collected from that subgroup alone, this proportion increases, making the model more likely to violate the fairness criteria, hence potentially disincentivizing additional data collection on that subgroup. Here we empirically examine this hypothesis.

\subsection{Study Description}
Using the same datasets, subgroups, pre-processing, cleaning, and models outlined in Appendix \ref{appendix:methods}, we calculate the value of the following expression from Equation \eqref{eq:Kearns}:  
\begin{equation}\label{eqn:Kearns_metric}
    \alpha(G) |m(G) - m(\cdot)|
\end{equation}

Recall that $\alpha(G)$ is the proportion of group $G$ within the total population, and that $m$ is some model performance metric (as in Appendix \ref{appendix:methods}, we use accuracy as our sample metric $m$ for this study).  The value $m(G)$ is the model performance metric evaluated only on subgroup $G$, while $m(\cdot)$ is the value of the model performance metric on the entire dataset.  

Kearns et al.\ \cite{Kearns_FairnessGerrymandering} use an auditing process wherein the value calculated from expression \eqref{eqn:Kearns_metric} must be below some threshold $\epsilon$ in order for a model to be considered fair.  Thus, we can think of expression \eqref{eqn:Kearns_metric} as describing \emph{unfairness} for group $G$.\footnote{All typical ways of defining ``fairness'' can be interpreted this way. A higher $\epsilon$ in \eqref{eq:typical} is interpreted as a \emph{de}crease in fairness and thus an increase in unfairness.}

\subsection{Methods}

We calculate the value in expression \eqref{eqn:Kearns_metric} on increasing subsamples of each dataset. We concentrate on small sugbroups $G$ that comprise no more than 10\% of the total population. Just as in the experiments described in Appendix \ref{appendix:methods}, we examine two subsampling scenarios: We subsample the subgroup in question only (to simulate gathering more subgroup data), and we subsample the entire dataset (to simulate gathering more population data).  Note that, of course, the values of $m$ depend on the model created, which depends on the subsample of the data used to create the model. 

Many of the datasets ({\tt heart\_disease, nursery, ricci, student\_math, student\_por, tae} and {\tt us\_crime}) don't have any subgroups comprising less than 10\% of the total population, and thus we exclude those datasets from this analysis.  From the other datasets, we concentrate on four: the {\tt adult, bank, meps20}, and {\tt titanic} datasets. The results for all other datasets are similar.  

In Section \ref{sec:incentive_compatibility}, we hypothesize that the protocol of Kearns et al.\ violates \emph{Incentive Compatibility}. Specifically, we hypothesize that when subsampling only small groups, the \emph{unfairness} value of expression \eqref{eqn:Kearns_metric} would \emph{increase}. Since the value $\alpha(G)$ does not change significantly when subsampling the entire population, we do not expect expression \eqref{eqn:Kearns_metric} to change much when subsampling the entire dataset, aside from the fact that potentially a better model might make expression \eqref{eqn:Kearns_metric} decrease.

\subsection{Results and Discussion}

\begin{figure}[h]
    \centering
    \includegraphics[width=\linewidth]{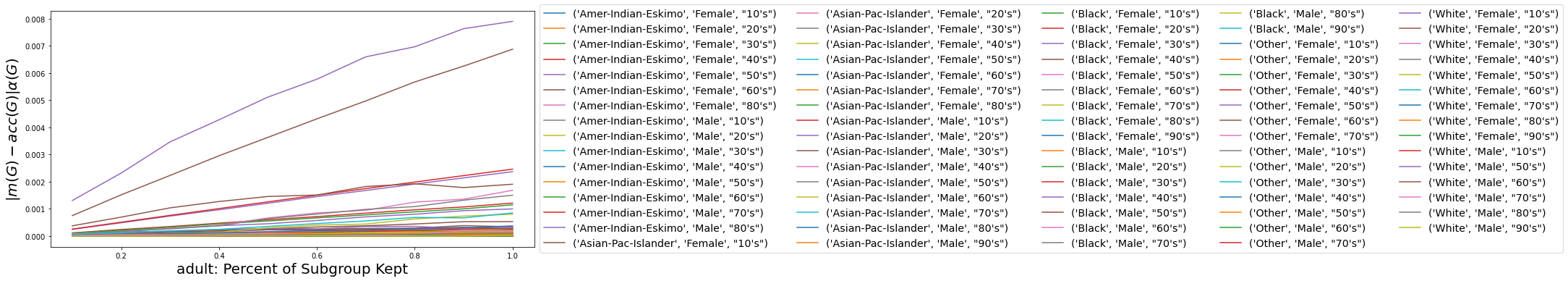}
    \includegraphics[width=\linewidth]{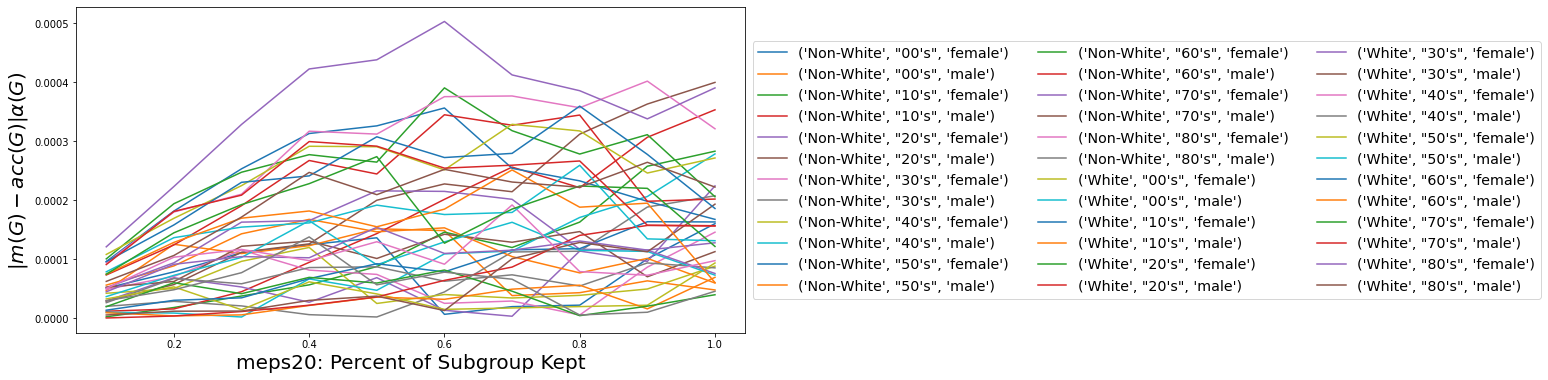}
    \includegraphics[width=0.49\linewidth]{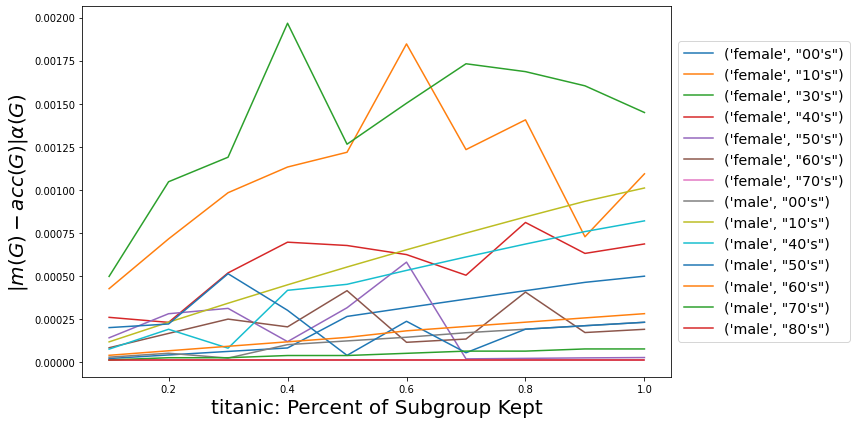}
    \includegraphics[width=0.49\linewidth]{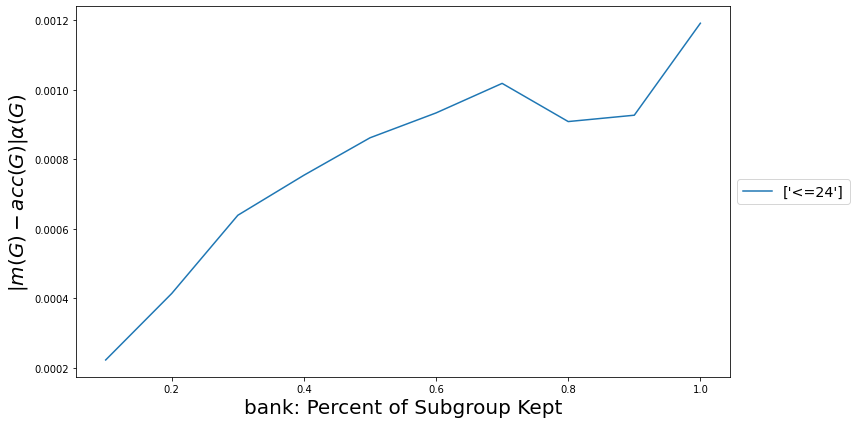}
    \caption{Values of expression \eqref{eqn:Kearns_metric} on the {\tt adult, meps20, titanic}, and {\tt bank} datasets.  Horizontal axis is the percent of the subgroup, vertical axis is unfairness (i.e., the value of expression \eqref{eqn:Kearns_metric}).}
    \label{fig:Kearns_subsample_subgroup}
\end{figure}

\begin{figure}[h]
    \centering
    \includegraphics[width=\linewidth]{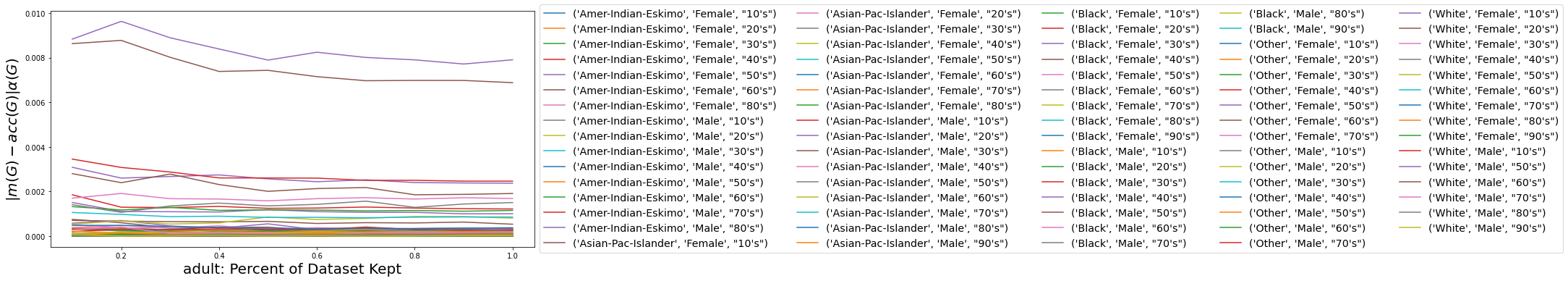}
    \includegraphics[width=\linewidth]{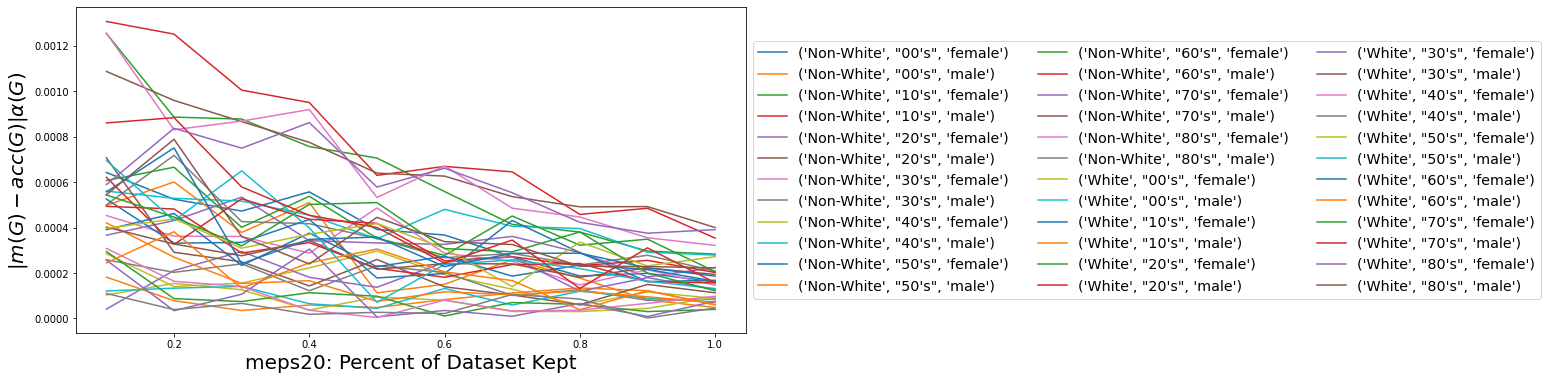}
    \includegraphics[width=0.49\linewidth]{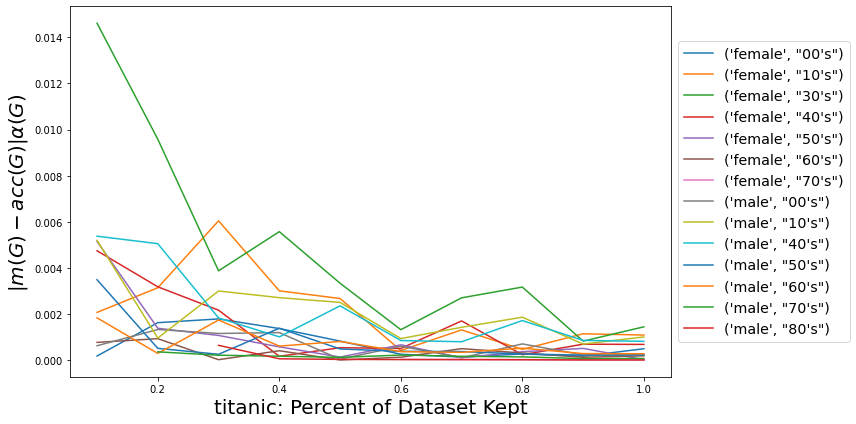}
    \includegraphics[width=0.49\linewidth]{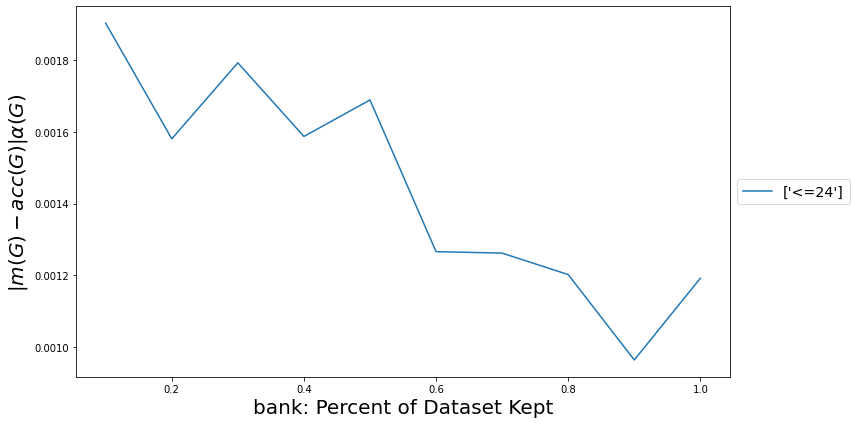}
    \caption{Values of expression \eqref{eqn:Kearns_metric} on the {\tt adult, meps20, titanic}, and {\tt bank} datasets.  Horizontal axis is the percent of the entire dataset kept, vertical axis is unfairness (i.e., the value of expression \eqref{eqn:Kearns_metric}).}
    \label{fig:Kearns_subsample_all}
\end{figure}

The results of subsampling just the small subgroup can be found in Figure \ref{fig:Kearns_subsample_subgroup}, and the results of subsampling the entire dataset can be found in Figure \ref{fig:Kearns_subsample_all}.

When only the small subgroup is subsampled (as in Figure \ref{fig:Kearns_subsample_subgroup}), we see the value of \eqref{eqn:Kearns_metric} increasing for all subgroups in the {\tt adult} and {\tt bank} datasets.  The picture is slightly more muddled in the {\tt meps20} and {\tt titanic} datasets, but these still show either a consistent increase or an initial increase for nearly all of the small subgroups.  In other words, the value \eqref{eqn:Kearns_metric} of unfairness \emph{increases}, indicating that the Kearns et al.\ auditing process discourages additional data collection of small subgroups, and thus violates \emph{Incentive Compatibility}.

When the entire dataset is subsampled (as in Figure \ref{fig:Kearns_subsample_all}), values of \eqref{eqn:Kearns_metric} remain remarkably consistent in the {\tt adult} dataset, and tend to decrease in the {\tt bank, meps20}, and {\tt titanic} datasets.  We can thus conclude that the Kearns et al.\ approach, while it discourages collecting additional data from only the smallest subgroups in a dataset (thereby not satisfying \emph{Incentive Compatibility}), does not appear to discourage additional data collection when each subgroup's proportion within the population stays consistent.

\end{document}